\documentclass[letterpaper]{article} 
\usepackage{aaai24}  
\usepackage{times}  
\usepackage{color}
\usepackage{svg}
\usepackage{amssymb}
\usepackage{amsmath}
\usepackage[round, sort]{natbib}
\usepackage{helvet}  
\usepackage{courier}  
\usepackage[hyphens]{url}  
\usepackage{graphicx} 
\usepackage{natbib}  
\usepackage{caption} 
\usepackage{algorithm}
\usepackage{algorithmic}

\usepackage{newfloat}
\usepackage{listings}
\DeclareCaptionStyle{ruled}{labelfont=normalfont,labelsep=colon,strut=off} 
\floatstyle{ruled}
\newfloat{listing}{tb}{lst}{}
\floatname{listing}{Listing}
\title{Spatial-Contextual Discrepancy Information Compensation \\for GAN Inversion}
\author{
    Ziqiang Zhang\textsuperscript{\rm 1},
    Yan Yan\textsuperscript{\rm 1}\thanks{Corresponding author (email: {yanyan@xmu.edu.cn}).},
    Jing-Hao Xue\textsuperscript{\rm 2},
    Hanzi Wang\textsuperscript{\rm 1}
}
\affiliations{
    \textsuperscript{\rm 1}Xiamen University, China\quad
    \textsuperscript{\rm 2}University College London, UK
}

\usepackage{bibentry}

\begin{document}

\maketitle

\begin{abstract}
	Most existing GAN inversion methods either achieve accurate reconstruction but lack editability or offer strong editability at the cost of fidelity. 
	Hence, how to balance the distortion-editability trade-off is a significant challenge for GAN inversion.  To address this challenge, we introduce a novel spatial-contextual discrepancy information compensation-based GAN-inversion method (SDIC), which
consists of a discrepancy information prediction network
(DIPN) and a discrepancy information compensation network
(DICN). SDIC follows a ``compensate-and-edit'' paradigm and successfully bridges the gap in image
details between the original image and the reconstructed/edited image. On the one hand, DIPN  encodes the multi-level spatial-contextual
information of the original and initial reconstructed
images and then predicts a spatial-contextual guided
discrepancy map with two hourglass modules. 
In this way, a reliable discrepancy map that models the contextual relationship and
captures fine-grained image details is learned. 
On the other hand, DICN incorporates the predicted
discrepancy information into both the latent code
and the GAN generator with different transformations, generating high-quality reconstructed/edited images. This effectively compensates for
the loss of image details during GAN inversion. Both quantitative and qualitative experiments demonstrate that
	our proposed method achieves the excellent distortion-editability trade-off at a fast inference speed for both image inversion and editing tasks. Our code is available at https://github.com/ZzqLKED/SDIC.
\end{abstract}

\section{Introduction}
Over the past few years, 
a variety of powerful GAN models, such as {PGGAN \cite{karras2017progressive} and}
StyleGAN \cite{karras2019style,karras2020analyzing}, have been developed to generate 
high-quality images based on the latent code 
in the latent space. 
Based on these models, we can manipulate some attributes of generated images by modifying the latent code. However, such manipulation is only applicable to the images given by the GAN
generator due to the lack of inference capability in GANs \cite{xia2022gan}.

Recently, GAN inversion methods \cite{xia2022gan} have been proposed to manipulate real images. These methods usually follow an ``invert first, edit later'' procedure, 
which first inverts the real image back into {a latent code of a pre-trained GAN model}, and then a new image can be reconstructed or edited from the inverted latent code.

Some GAN inversion methods \cite{richardson2021encoding,tov2021designing} invert a real image into {the native latent space of StyleGAN (i.e., the $\mathcal{W}$ space)} and achieve good editability. However, such a way inevitably leads to the loss of image details during inversion. As a result, 
the reconstructed images are often less faithful than the original images.  
Although some methods \cite{abdal2019image2stylegan,kang2021gan} extend the $\mathcal{W}$ space to the $\mathcal{W}^{+}$/$\mathcal{W}$* space or perform per-image optimization to enhance the reconstruction fidelity, their editability can be greatly affected. 
Such a phenomenon is also known as 
the distortion-editability {trade-off}, indicating the conflict between image reconstruction fidelity and image editing quality.

To alleviate the distortion-editability {trade-off}, 
some recent methods (such as HFGI \cite{wang2022high} and CLCAE \cite{liu2023delving}) recover the missing information
by enriching the latent code and the feature representations of a particular layer in the GAN generator. 
In this way, they can achieve better fidelity than traditional GAN inversion methods. 
However,  HFGI considers the distortion information only at the pixel level, which easily introduces significant artifacts; CLCAE generates images based on {contrastive learning}, and it may still lose some image details and degrade the editability. Therefore, existing GAN inversion methods still suffer from the gap in image details between the original
image and the reconstructed/edited image. 

To address the above problems, we propose a novel spatial-contextual discrepancy information compensation-based GAN inversion method (SDIC), which consists of a discrepancy information prediction network (DIPN)  and a discrepancy information compensation network (DICN). 
SDIC adopts a ``compensate-and-edit'' paradigm, which first compensates both the latent code and the GAN generator with the spatial-contextual guided discrepancy map,
and then performs image inversion or editing. 

Specifically, DIPN, which consists of a two-branch spatial-contextual hourglass module and a discrepancy map learning hourglass module, is designed to encode the multi-level spatial-contextual information of the original and initial reconstructed images, and 
predict a spatial-contextual guided discrepancy map. 
In DIPN, a spatial attention mechanism is leveraged to enable the network to adaptively select important parts for feature fusion. As a result, DIPN can accurately learn a reliable discrepancy map, which effectively captures the contextual relationship and fine-grained image details. 
Then, DICN is introduced to incorporate the discrepancy information into both the latent code and the GAN generator, generating high-quality reconstructed/edited images.

In summary, our main contributions are given as follows:
\begin{itemize}
	\item 
     We propose a novel GAN inversion method, which successfully exploits the multi-level spatial-contextual information of the original image to compensate for the missing information during inversion.  Based on the ``compensate-and-edit'' paradigm, our method can generate high-quality and natural reconstructed/edited images containing image details without introducing artifacts.
	\item We design DIPN to accurately predict the spatial-contextual guided discrepancy map with two hourglass modules and DICN to leverage this map to effectively compensate for the information loss in both the latent code and the GAN generator with different transformations. Therefore, our method can achieve an excellent distortion-editability trade-off. 
	\item We perform qualitative and quantitative experiments 
	to validate the superiority of our method in fidelity and editability against state-of-the-art GAN 
	inversion methods.
\end{itemize}

\section{Related Work}
\noindent \textbf{GAN Inversion.} 
Existing GAN inversion methods can be divided into three categories: optimization-based, encoder-based, and hybrid methods. 
Optimization-based methods \cite{abdal2020image2stylegan++,bau2020semantic,gu2020image,zhu2020improved} directly optimize the latent code or the parameters of GAN to minimize the reconstruction error for each given image. 
Although these methods can reconstruct high-fidelity images, they usually suffer from high computational cost and poor editability. 
Encoder-based methods \cite{alaluf2021restyle,hu2022style,kang2021gan,tov2021designing} train an encoder to learn the mapping from a given image to a latent code and perform some operations on the latent code. 
Compared with the optimization-based methods, the encoder-based methods show better editability at a faster inference speed, but their reconstruction quality is much lower. 
Hybrid methods \cite{zhu2020domain,bau2019seeing} leverage an encoder to learn a latent code and then optimize the obtained latent code. They can balance the inference time and reconstruction quality. However, it is not trivial to determine the exact latent code. Recently, some methods \cite{alaluf2022hyperstyle,dinh2022hyperinverter} introduce the hypernetwork to iteratively optimize the parameters of the GAN generator, obtaining better reconstruction results. 

Our method belongs to the encoder-based methods. Conventional encoder-based methods (such as pSp \cite{chang2018pyramid} and e4e \cite{tov2021designing}) extend the $\mathcal{W}$ space to the $\mathcal{W}^{+}$/$\mathcal{W}$* space to improve the reconstruction fidelity. However, 
the low-dimensional latent code still limits the reconstruction performance.  
Moreover, the editing flexibility of the  $\mathcal{W}^{+}$/$\mathcal{W}$* space is reduced. 
In contrast, we propose to compensate the latent code and the GAN generator by exploiting the multi-level spatial-contextual information.  
In this way, the expressiveness of the latent code and generator is largely improved, alleviating the information loss due to inversion.
Meanwhile, we explicitly control the proximity of the latent code to the $\mathcal{W}$ space, providing better editability. 

\noindent \textbf{Latent Space Editing.} The latent spaces of pre-trained StyleGAN generators show good semantic interpretability, which enables diverse and flexible image editing. A number of methods 
have been developed to identify meaningful semantic directions for the latent code. Supervised methods \cite{abdal2021styleflow,hutchinson2019detecting,goetschalckx2019ganalyze,shen2020interpreting} require pre-trained attribute classifiers or data with attribute annotations. InterFaceGAN \cite{shen2020interpreting} employs a support vector machine to identify the hyperplane that splits two binary attributes and considers the normal of the hyperplane as the manipulation direction. Unsupervised methods  \cite{harkonen2020ganspace,shen2021closed,voynov2020unsupervised,wang2021geometry} can discover unknown manipulation directions but require manual labeling of the found operational directions.   GANspace \cite{harkonen2020ganspace} discovers multiple manipulation directions using principal component analysis. 
In this paper, we employ InterFaceGAN and GANspace for latent space editing due to their good editing performance. 
\begin{figure*}[!t] 
	\centering 
	\includegraphics[width=0.85\textwidth]{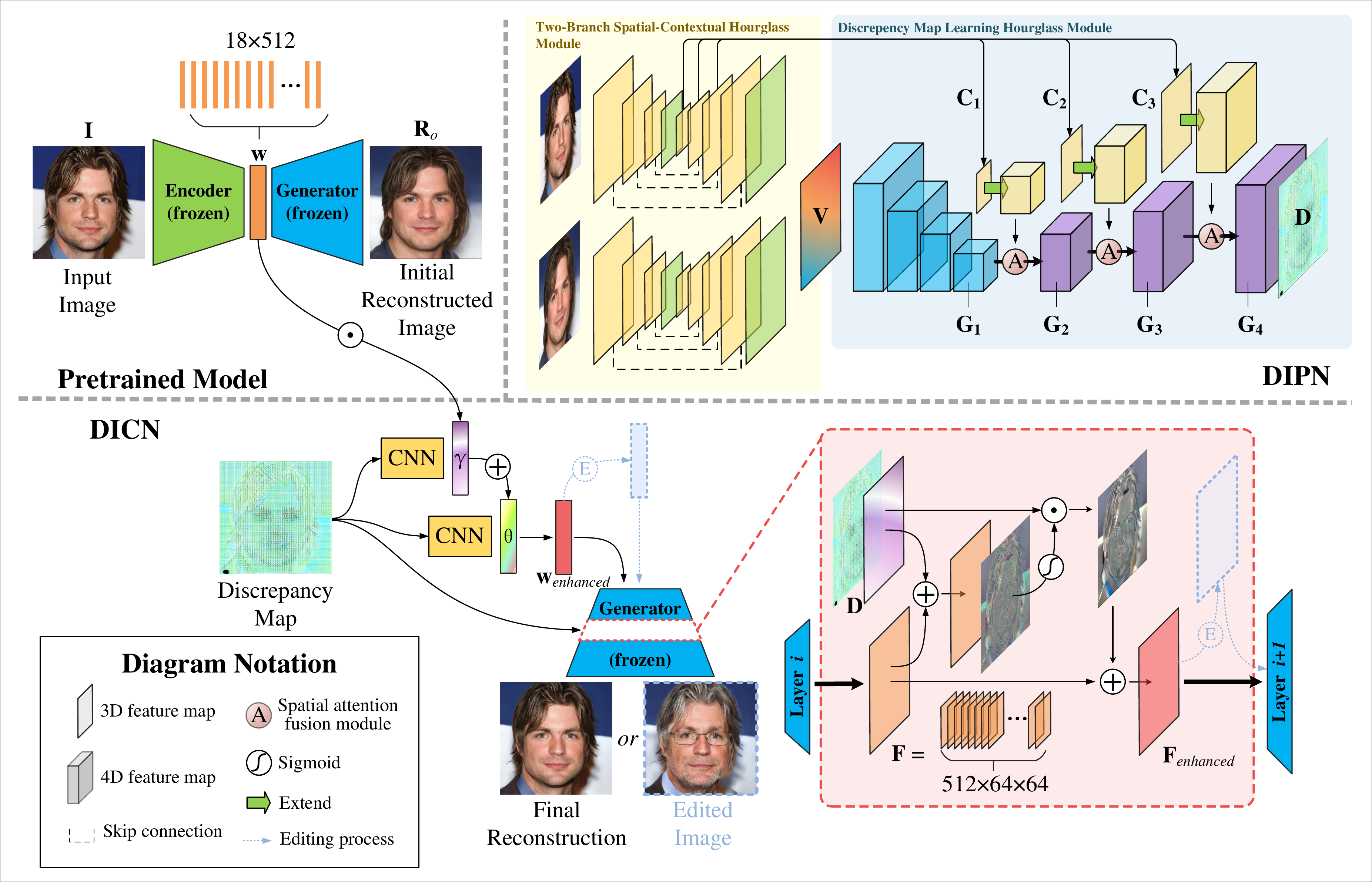} 
	 
	\caption{
		The architecture of SDIC, which consists of DIPN and DICN. DIPN contains a two-branch spatial-contextual hourglass module and
		a discrepancy map learning hourglass module. First, the original image $\mathbf{I}$ and the initial reconstructed image $\mathbf{R}_{o}$ (obtained by a pre-trained e4e model) are fed into DIPN  to predict the discrepancy map. 
		Then, the discrepancy map is fed into  DICN  for feature compensation in both the latent code and the GAN generator. }

	
\label{Fig.1} 
\end{figure*}

\noindent \textbf{Spatial-Contextual Information.} 
Convolutional neural network (CNN) encodes both low-level features at the early stages and high-level features at the later stages. The low-level features are rich in spatial details while the high-level features capture the contextual information, which encodes the visual knowledge from an object/patch and its surrounding backgrounds/patches. The spatial-contextual information plays an important role in many computer vision tasks, such as object detection and semantic segmentation.  
Some methods \cite{choi2010exploiting,li2016attentive} exploit the contextual information between the object and its surrounding background to improve the object detection performance. A few works  \cite{chang2018pyramid} improve the quality of the detailed parts of the disparity map by exploiting multi-scale spatial-contextual information.  

Unfortunately, the spatial-contextual information is not well exploited in exisiting GAN inversion methods. In this paper, we introduce the spatial-contextual information (obtained from the original image) to the discrepancy map prediction between the original image and the initial reconstructed image. In this way,
our method can preserve more appearance details and generate clearer edges, greatly reducing 
the artifacts of the reconstructed/edited images.   

\section{Proposed Method}
\subsection{Overview}
\noindent \textbf{Motivation.} 
Conventional GAN inversion methods \cite{alaluf2021restyle,collins2020editing,kang2021gan,pidhorskyi2020adversarial,tov2021designing} invert a real image into the latent space of a pretrained GAN model and attain good editability. However, they usually suffer from the low fidelity of generated images due to severe information loss during the inversion process. 
To deal with this, some recent methods (such as HFGI) follow an ``edit-and-compensate'' paradigm, which computes the distortion map (between the original image and the initial edited image), and then encodes the distortion map to obtain the latent map. 
In this way, the latent map can be combined with the latent code to compensate for the loss of image details.  
However, the images generated by these methods easily suffer from artifacts. 

The problem of artifacts can be ascribed to the fact that the latent map only encodes the pixel-level spatial information (note that the distortion map is computed by {subtracting the initial edited image from the original image}). As a result, the latent map ignores high-level contextual information (i.e., the relationship between individual pixels and their surrounding pixels) and involves some attribute-specific disturbance caused by the adoption of the initial edited image.




\noindent \textbf{Design.} To address the above problems, we exploit both the spatial and contextual information of the original image. This enables the network to learn the contextual relationship between pixels and spatial details, significantly reducing the artifacts. To this end, we propose a novel SDIC method. Instead of using the ``edit-and-compensate'' paradigm in previous methods, SDIC adopts a ``compensate-and-edit'' paradigm, which enables high-quality and flexible editing of attributes. Notably, SDIC 
effectively exploits the spatial-contextual guided discrepancy information between the original image and the initial reconstructed image and leverages this information to compensate both the latent code and the GAN generator. 
In this way, our method achieves a good distortion-editability trade-off. 

The network architecture of SDIC is given in Fig.~\ref{Fig.1}. SDIC consists of a discrepancy information
prediction network (DIPN) and a discrepancy information compensation network (DICN). 
Given the original image and the initial reconstructed image,  DIPN (which contains a two-branch spatial-contextual hourglass module and a discrepancy map learning hourglass module) predicts a discrepancy map. In particular, we incorporate 
the spatial and contextual information of the original image into the different layers of the discrepancy map learning hourglass module. As a result, the discrepancy map not only encodes fine-grained image
details but also models the contextual relationship. The discrepancy map is subsequently fed into  DICN to perform feature compensation for the information loss in both the latent code and the GAN generator, obtaining the enhanced latent code and the enhanced latent map. 



The attribute editing takes a similar process as the inversion process
except that the enhanced latent code and enhanced latent map are modified by attribute editing operations in DICN.  



%

\subsection{Discrepancy Information Prediction Network} 

\subsubsection{Two-Branch Spatial-Contextual Hourglass Module.}
The two-branch 
spatial-contextual hourglass module  takes both the original image $\mathbf{I}$ $\in$ $\mathbb{R}^{3\times H \times W}$ and the initial reconstructed image $\mathbf{R}_{o}$ $\in$ $\mathbb{R}^{3\times H \times W}$ generated from the pre-trained model (we use the popular e4e model \cite{tov2021designing}) as inputs and extracts their spatial and contextual information, where $H$ and $W$ denote the height and width of the image, respectively. 
Generally, this module consists of two parallel hourglass branches, where two branches have the same structures.  
Each branch consists of a convolutional block and a U-Net style upsampling block \cite{ronneberger2015u}. 

Specifically, given an input image ($\mathbf{I}$ or $\mathbf{R}_o$), it is first fed into a convolutional block (consisting of five 3$\times$3 convolutional layers). In the convolutional block, the second, fourth, and fifth layers perform convolution with stride 2, reducing the feature map size to 1/2, 1/4, and 1/8 of the original size, respectively. Such a way gradually expands the receptive fields and captures coarse-to-fine information at different scales. 

Next, the reduced feature map is fed into a U-Net style upsampling block with skip-connection at different scales.
Technically, the upsampling block contains three 4$\times$4 convolutional layers with stride 1. Each convolutional layer is preceded by an upsampling layer that uses nearest-neighbor interpolation to upsample the feature map.
By doing so, we can repeatedly upsample the feature map until its size reaches $3 \times H \times W$. In this way, the sizes of the feature map are gradually resized to 1/4 and 1/2 of the input image size. 
Meanwhile, a 3$\times$3 convolutional layer is also applied to merge the skip connection and the final upsampled feature.  The receptive fields are gradually reduced during upsampling. Thus, more fine-grained information can be obtained. Note that during upsampling, the three feature maps {$\mathbf{C}_1, \mathbf{C}_2, \mathbf{C}_3$}, whose sizes correspond to {1/8,} 1/4, and 1/2 of the original image size, respectively,  serve as the multi-resolution feature maps for the subsequent discrepancy map learning 
 hourglass module. These feature maps model the multi-level spatial-contextual information at different resolutions. Generally, the higher-resolution feature map 
captures more spatial details while lower-resolution feature map encodes more contextual information.  


Finally, the module outputs two feature maps with the size of $C \times H \times W$ ($C$=48) corresponding to the original image and the initial reconstructed image, respectively.



\subsubsection{Discrepancy Map Learning Hourglass Module.}
The two feature maps from the two-branch spatial-contextual hourglass module are concatenated together to obtain the feature volume $\mathbf{V} \in \mathbb{R}^{2C \times H \times W}$, which is taken as the input of the discrepancy map hourglass learning module. This module consists of a downsampling block and a fusion block. Based on the feature volume $\mathbf{V}$, we apply a downsampling block to obtain the initial discrepancy information. Then, the fusion block combines the initial discrepancy information with the spatial-contextual information of the original image from the two-branch spatial-contextual hourglass module to predict the discrepancy map.

Specifically, the feature volume is first fed into the downsampling block, which uses three downsampling layers to increase the receptive fields. 
Each downsampling layer consists of a 3$\times$3 $\times$3 3D convolution with stride 2 and a 3 $\times$3$\times$3 3D convolution  with stride 1. As a result, we get the discrepancy features  $\mathbf{G}_1 \in \mathbb{R}^{C \times 48 \times H/8 \times W/8}$ after downsampling. $\mathbf{G}_1$ is then fed into a fusion layer (consisting of a spatial attention fusion module and an upsampling layer) to obtain the spatial-contextual guided discrepancy feature $\mathbf{G}_2$ at a larger resolution. Each upsampling layer consists of a 4$\times$4$\times$4 3D transposed convolution with stride 2 and two 3$\times$3$\times$3 3D convolutions with stride 1. Similar operations are repeated to obtain higher resolution features $\mathbf{G}_3$ and $\mathbf{G}_4$. 

\begin{figure}[!t] 
\centering 
\includegraphics[width=0.4\textwidth]{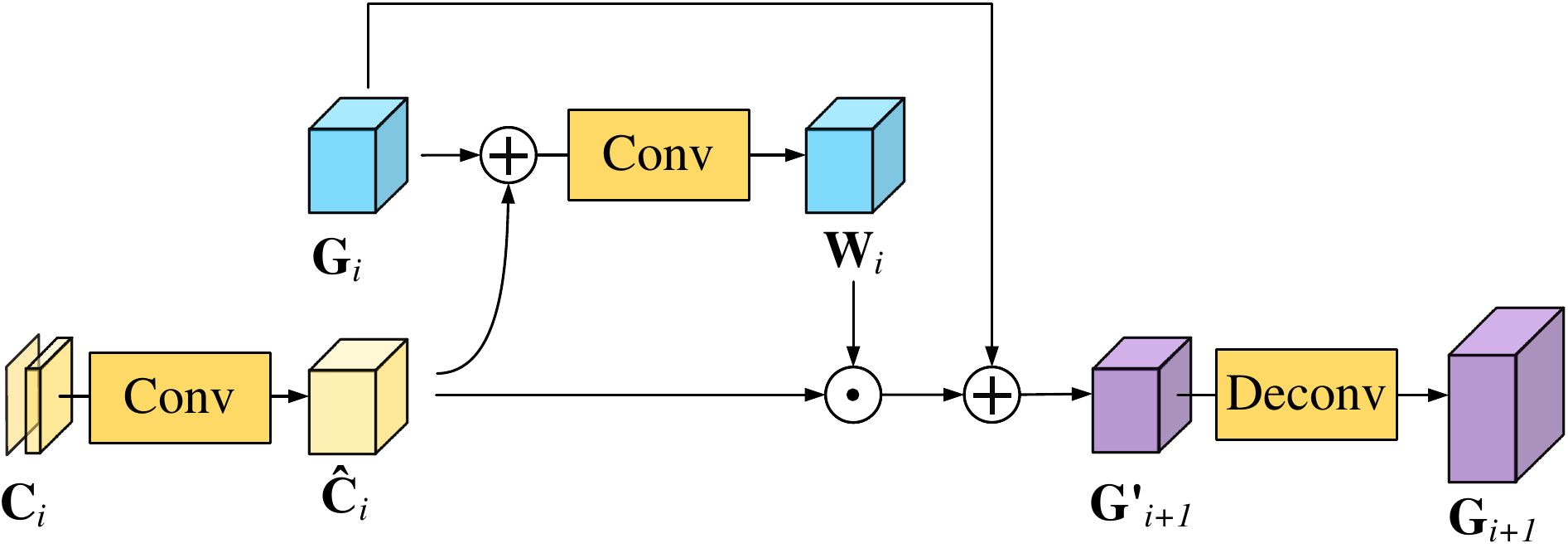} 
 
\caption{The architecture of the spatial attention fusion module.}

\label{Fig.2} 
\end{figure}

Instead of adding or concatenating the discrepancy feature and the spatial-contextual feature, we incorporate a spatial attention fusion module \cite{woo2018cbam}. This module enables us to adaptively select important regions of features for fusion. The architecture of the spatial attention fusion module is given in Fig.~\ref{Fig.2}. 

Technically, $\mathbf{C}_i$  ($i$=1, 2, 3) is first upsampled by a 3D convolutional layer to obtain $\widehat{\mathbf{C}}_i$ which has the same dimensions as $\mathbf{G}_i$. Then, we add ${\mathbf{\widehat{C}}}_{i}$ and $\mathbf{G}_{i}$ to obtain the enhanced feature, which  is fed into a 5$\times$5$\times$5 3D convolutional layer to get spatial attention weights $\mathbf{W}_i$, i.e., 
\begin{equation}
\mathbf{W}_i=\sigma(f^{5 \times 5 \times5}(\mathbf{G}_i+\widehat{\mathbf{C}}_i)),
\end{equation}
where $\sigma$ denotes the Sigmoid function and $f^{5 \times 5 \times5}$ denotes the 5$\times$5$\times$5 3D convolution operation. 

The attention weights reflect the importance of each spatial location that needs to be emphasized. Hence, we fuse the discrepancy feature and the spatial-contextual feature as
\begin{equation}
\mathbf{G}_{i+1}^{'}=\mathbf{W}_i \odot \widehat{\mathbf{C}}_i + \mathbf{G}_i,
\end{equation}
where 
`$\odot$' denotes the Hadamard product. 

Next, $\mathbf{G}_{i+1}^{'}$
is fed into 
 a 5$\times$5$\times$5 3D deconvolution layer to obtain $\mathbf{G}_{i+1}$. 
Finally, we get $\mathbf{G}_4 \in \mathbb{R}^{1 \times 3 \times H \times W}$ and then perform a 3$\times$3 convolutional operation on $\mathbf{G}_4$ to generate the final discrepancy map $\mathbf{D} \in \mathbb{R}^{3 \times H \times W}$.



\subsection{Discrepancy Information Compensation Network}
As we previously mentioned, {due to the information loss during inversion, the information involved in the latent code $\mathbf{w}$ (with the size of $18\times512$) is inadequate}. Therefore, many existing methods extract additional information to compensate $\mathbf{w}$. However, such a way cannot guarantee the preservation of image details
because of the low dimensionality of  $\mathbf{w}$. In this paper, we introduce to compensate for the information loss in both $\mathbf{w}$ and the early layer of the GAN generator. The architecture of DICN is illustrated in Fig.~\ref{Fig.1}.


On the one hand, to compensate the latent code $\mathbf{w}$ with the discrepancy map $\mathbf{D}$, we leverage the conventional linear affine transformation. As shown in Fig.~\ref{Fig.1}, we apply two 3$\times$3 convolutional layers to the discrepancy map obtained from DIPN,  predicting the scaling parameter $\gamma \in \mathbb{R}^{18 \times 512}$ and the displacement parameter $\theta\in \mathbb{R}^{18 \times 512}$, i.e., 
\begin{equation}
\begin{aligned}
\gamma=f^g(\mathbf{D}), \theta=f^t(\mathbf{D}),\\
\end{aligned}
\end{equation}
where $f^g$ and $f^t$ denote the convolution layers. 

Based on the above, we apply a channel scaling operation to $\mathbf{w}$ using the scaling parameter $\gamma$, followed by a channel displacement operation using the displacement parameter $\theta$. This process effectively filters out uninformative features while compensating for insufficient detailed features. The affine transformation expands the representation space of the generator and facilitates the extraction of high-fidelity features in StyleGAN. The above process is expressed as
\begin{equation}
\begin{aligned}
\mathbf{w}_{enhanced}(i)
=\gamma_{i} \odot \mathbf{w}_i+\theta_{i},\\	
\end{aligned}
\end{equation}
where $\mathbf{w}_{enhanced}(i)$ is the $i$-th row of the enhanced latent code $\mathbf{w}_{enhanced}$; $\mathbf{w}_i$ is the $i$-th row of $\mathbf{w}$; $\gamma_i$ and $\theta_i$ are the $i$-th rows of $\gamma$ and $\theta$, respectively. 


On the other hand, to compensate for the information loss in the generator, instead of using the affine transformation, 
we take a similar fusion way as done in DIPN. We apply the discrepancy map to compensate the output of the early layer 
of the generator (we choose layer 7 in this paper as suggested by HFGI). We denote the output of this layer as the latent map $\mathbf{F}$ (with the size of $512\times64\times64$). We adopt the spatial attention fusion module to adaptively select the important parts and 
suppress the unimportant parts of features. Then, we add the attention map to $\mathbf{F}$, that is, 
\begin{equation}
\begin{aligned}
\mathbf{F}_{enhanced}=\sigma(f^{c1}(\mathbf{F}+f^{c2}(\mathbf{D})))\odot f^{c2}(\mathbf{D})+\mathbf{F},
\end{aligned}
\label{eq:5}
\end{equation}
where $f^{c1}$ and $f^{c2}$ denote convolutional blocks ($f^{c1}$ contains two 3$\times$3 convolution layers and $f^{c2}$ contains four 3$\times$3 convolution layers). $\mathbf{F}_{enhanced}$ denotes the enhanced latent map which is further fed to {the next layer of the generator.}

{Note that some methods (such as FS {\cite{xuyao2022}}, CLCAE, and HFGI) also 
operate on both the latent code and the generator. However, the differences between these methods and our method are significant. 
FS and CLCAE  follow the ``compensate-and-edit'' paradigm, where they train an additional encoder to generate a new latent code while obtaining a new latent map to replace one layer of the generator. 
Although such a way improves the fidelity, the editability is affected (since the new latent space is greatly different from the $\mathcal{W}$ space that is proven to have excellent editability).
HFGI follows the  ``edit-and-compensate'' paradigm and leverages the pixel-level distortion map (from the original image and the initial edited image) to compensate the latent map by a linear transformation, limiting the editing performance.
Moreover, the above methods ignore spatial-contextual information, resulting in losing some image details or introducing artifacts.
In contrast, we generate a spatial-contextual guided discrepancy map (from the original image and the initial reconstructed image) to compensate the latent map by a nonlinear transformation, adaptively fusing features for better compensation and expanding the representation space.



\subsection{Attribute Editing Operations}
We perform operations on both the enhanced latent code and the enhanced latent map for attribute editing. To be specific, the enhanced latent code $\mathbf{w}_{enhanced}$ is first modified by the mainstream latent space editing method \cite{shen2020interpreting,harkonen2020ganspace} to obtain the edited latent code (used as the input of the GAN generator). Then, the enhanced latent  {map} is modified as follows. We first obtain the initial reconstructed image $\mathbf{R}_o$ and the initial edited image $\mathbf{E}_o$. Then the latent map $\mathbf{F}^R$ of the reconstructed image and the latent map $\mathbf{F}^E$ of the edited image at layer 7 of the generator are extracted. 
Assume that the enhanced latent {maps} of $\mathbf{F}^R$ and $\mathbf{F}^E$ are represented as 
$\mathbf{F}_{enhanced}^E$ and $\mathbf{F}_{enhanced}^R$, respectively. 
During the attribute editing, we expect that 
the difference between $\mathbf{F}_{enhanced}^R$ and $\mathbf{F}_{enhanced}^E$  should be close to that 
between $\mathbf{F}^R$ and $\mathbf{F}^E$ in the latent space 
to ensure editability. Hence, instead of generating $\mathbf{F}_{enhanced}^E$ via Eq.~(\ref{eq:5}), we add the difference between $\mathbf{F}^R$ and $\mathbf{F}^E$ to $\mathbf{F}_{enhanced}^R$ for predicting $\mathbf{F}_{enhanced}^E$, that is,
\begin{equation}	\mathbf{F}_{enhanced}^{E}=\mathbf{F}_{enhanced}^R+\mathbf{F}^{E}-\mathbf{F}^R,
\end{equation}
where $\mathbf{F}_{enhanced}^R$ is obtained via Eq.~(\ref{eq:5}).

\subsection{Joint Loss}
During the training stage, the parameters of the generator are frozen so that we can focus on optimizing the encoding process. We design a joint loss to achieve high reconstruction quality and good editability.

For the reconstruction quality, we define the reconstruction loss for the original image $\mathbf{I}$ and the reconstructed image $\mathbf{R}_f$ as
\begin{equation}	\mathcal{L}_{rec}=\mathcal{L}_2(\mathbf{I},\mathbf{R}_f)+\lambda_{LPIPS} \mathcal{L}_{LPIPS}(\mathbf{I},\mathbf{R}_f)+\lambda_{ID} \mathcal{L}_{ID}(\mathbf{I},\mathbf{R}_f),
\end{equation}
{where $\mathcal{L}_2(\mathbf{I},\mathbf{R}_f)$ 
denotes the Euclidean distance between $\mathbf{I}$ and $\mathbf{R}_f$ to evaluate the structural similarity; $\mathcal{L}_{L P I P S}(\mathbf{I},\mathbf{R}_f)$ denotes the LPIPS loss \cite{zhang2018unreasonable}  to evaluate the perceptual similarity;  $\mathcal{L}_{ID}=1-<\textrm{F}(\mathbf{I}),\textrm{F}(\mathbf{R}_f)>$ explicitly encourages the encoder to minimize the cosine similarity between $\mathbf{I}$ and $\mathbf{R}_f$, which can measure the identity consistency. Here, $\textrm{F}(\cdot)$ represents the feature extractor (we use the pre-trained ArcFace model \cite{deng2019arcface}  for the face domain and the pre-trained ResNet-50  model \cite{tov2021designing} for other domains). $\lambda_{LPIPS}$ and $\lambda_{ID}$ indicate the balancing weights.}

{To ensure the editability, we also incorporate the editing loss,  which is defined as}
\begin{equation}
\mathcal{L}_{edit}=\mathcal{L}_{1}(\mathbf{w},\mathbf{w}_{enhanced})+\mathcal{L}_{1}(\mathbf{F},\mathbf{F}_{enhanced}),
\end{equation}
{where $\mathcal{L}_{1}(\cdot,\cdot)$ denotes the $L_1$ norm.
This loss is leveraged to constrain the distances between $\mathbf{w}$ and $\mathbf{w}_{enhanced}$ as well as those between $\mathbf{F}$ and $\mathbf{F}_{enhanced}$. In this way, we can keep the latent codes close to the $\mathcal{W}$ space, beneficial to maintain the editability, as suggested in \cite{tov2021designing}.}

{Finally, the joint loss is expressed as}
\begin{equation}
\mathcal{L}_{joint}=\mathcal{L}_{rec}+\lambda_{edit} \mathcal{L}_{edit},
\end{equation}
where $\lambda_{edit}$ indicates the balancing weight. 

\begin{figure} [!t]
	\centering 
	\includegraphics[width=0.5\textwidth]{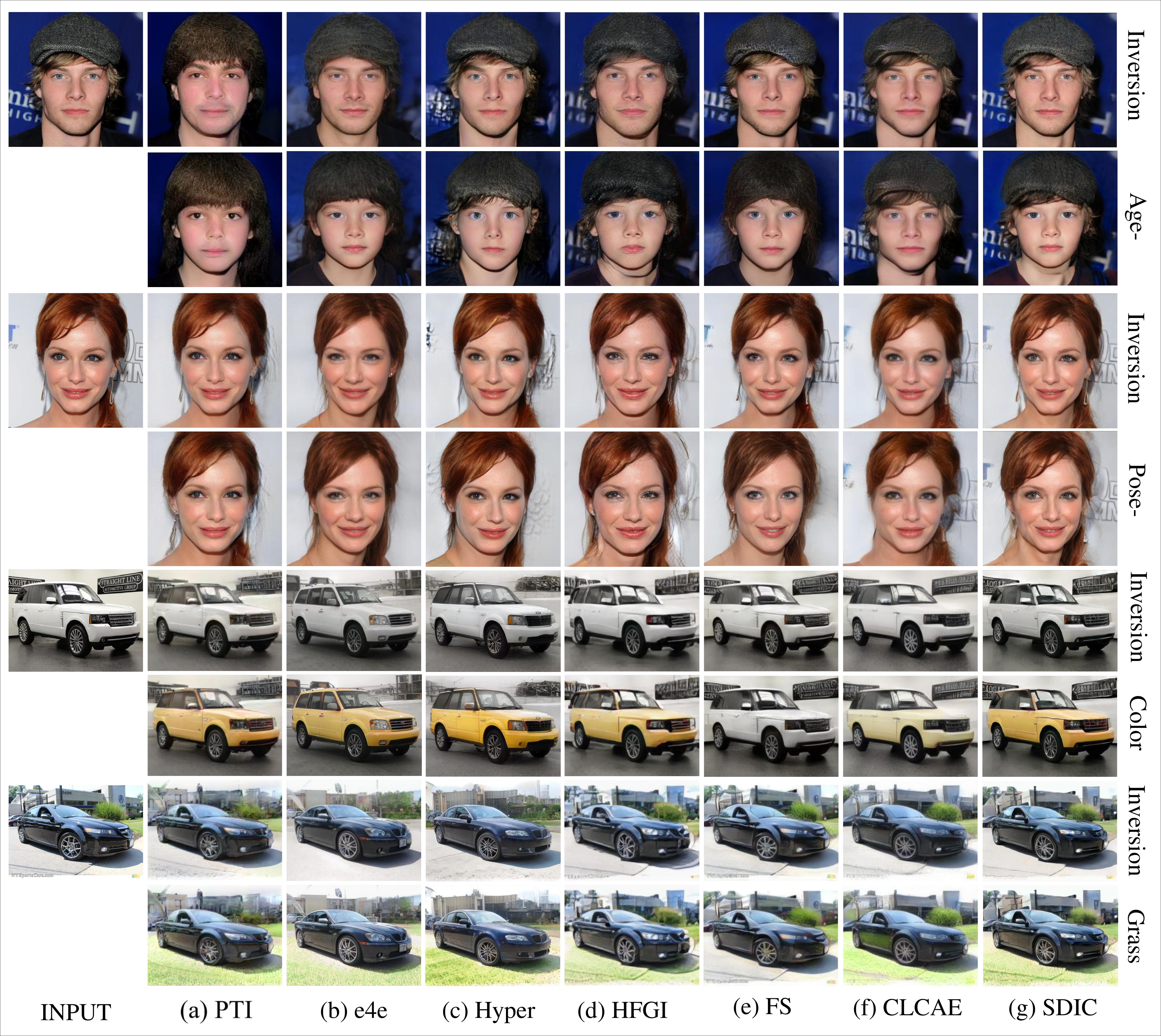} 
	 
	\caption{Qualitative comparison between SDIC and several state-of-the-art methods on image inversion and editing tasks.  
 More results are shown in \textit{Supplement B}.} 
	\label{Fig.3} 
\end{figure}

\section{Experiments}
\subsection{Experimental Settings}
\noindent \textbf{Datesets.} We evaluate our method on two domains: human faces and cars. For the face domain, we adopt the widely-used FFHQ dataset \cite{karras2019style} for training and the CelebA-HQ dataset \cite{karras2017progressive,liu2015deep} for testing. For the car domain, we used the Stanford car dataset \cite{krause20133d} for training and testing. 

\begin{table*}[!t]
	\begin{center}
		\renewcommand\arraystretch{0.9}
		\begin{tabular}{l| c c c c c c | c c}

			\hline
			&\multicolumn{6}{|c|}{Inversion} &\multicolumn{2}{c}{Editing}\\
			\hline
			Method & $ID\uparrow$ & $SSIM\uparrow$ & $PSNR\uparrow$ & $LPIPS\downarrow$ & $L_2~(MSE)\downarrow$ & $Time~(s)\downarrow$ & $ID\uparrow$ & $User~ Study\uparrow$\\
			\hline
			PTI & 0.832 & 0.703 & 24.355 & 0.110 & 0.012 & 201.357 & \textbf{0.726} & 22.500\%\\
			\hline
			e4e & 0.495 & 0.537 & 19.390 & 0.206 & 0.050 & \textbf{0.033} & 0.452 & 18.750\%\\
			HyperStyle & 0.737 & 0.624 & 22.513 & 0.104 & 0.025 & 0.710 & 0.663 & 21.668\%\\
			HFGI & 0.606 & 0.641 & 22.372 & 0.136 & 0.026 & 0.108 & 0.610 & 22.918\%\\
			FS & 0.815 & 0.648 & 23.797 & 0.934 & 0.015 & 0.568 & 0.492 & 17.083\%\\
			CLCAE & 0.708 & 0.725 & 25.665 & 0.083 & 0.012 & 0.125 & 0.518 & 22.918\%\\
			\hline
			SDIC & \textbf{0.871} & \textbf{0.815} & \textbf{27.672} & \textbf{0.057} & \textbf{0.007} & 0.321 & \textbf{0.726} & \textbf{65.832}\%\\
			\hline
		\end{tabular}
		\caption{Quantitative comparison for inversion/editing quality on the face domain. {The value of \textit{User Study} is the percentage of users that choose this method.} The best results are in bold.}		 
		\label{tab1} 
		
	\end{center}
\end{table*}

\noindent \textbf{Comparison Methods.} We compare our SDIC method with various GAN inversion methods, including optimization-based method PTI \cite{roich2022pivotal} and encoder-based methods (e4e \cite{tov2021designing}, HyperStyle \cite{alaluf2022hyperstyle}, HFGI \cite{wang2022high}, FS \cite{xuyao2022}, and CLCAE \cite{liu2023delving}). All the results of these comparison methods are
obtained by using the trained models officially provided by the corresponding authors. 

\noindent \textbf{Implementation Details.} 
In this paper, we adopt 
InterfaceGAN \cite{shen2020interpreting} for face image editing and
GANSpace \cite{harkonen2020ganspace}  for car image editing. We use the pre-trained StyleGAN generator and the e4e encoder in our method. 
The sizes of the input and output of the network are both 1024$\times$1024. $\lambda_{LPIPS}$, $\lambda_{ID}$, and $\lambda_{edit}$  are empirically set to 0.8, 0.2, and 0.5, respectively. We use the ranger optimizer \cite{yong2020gradient} with a learning rate of 0.001 and a batch size of 2. Our model is trained 100,000 steps on the NVIDIA GeForce RTX 3080 GPU.

\subsection{Reconstruction Results}
\noindent \textbf{Quantitative Evaluation.} 
We quantitatively compare our method with state-of-the-art GAN inversion methods. The results are shown in Table~\ref{tab1}. We use the $SSIM$, $PSNR$, and $L_2$ to measure the reconstruction error, and the $LPIPS$ \cite{zhang2018unreasonable} for the perceptual quality. We also use CurricularFace \cite{huang2020curricularface} to extract the features of two images and calculate their cosine similarity as the $ID$ distance, which can measure the identity similarity between each reconstruction image and the original input image.  These metrics are evaluated on the first 1,000 images of CelebA-HQ. In addition, we also report the inference time obtained by these methods.

As we can see, our method significantly outperforms both the encoder-based methods (HFGI, FS, CLCAE, and HyperStyle) and the optimization-based method {(PTI)} in terms of reconstruction quality (including $ID$, $SSIM$, $PSNR$, $LPIPS$, and $L_2$}). Notably, our method achieves a much faster inference speed than the optimization-based method.
In a word, SDIC obtains the highest fidelity at a fast inference speed among all the competing methods.

\noindent \textbf{Qualitative Evaluation.}
Fig. \ref{Fig.3} gives the qualitative comparison between SDIC and several state-of-the-art methods. 
{For the face domain, SDIC effectively preserves background and foreground details. In the first row of Fig.~\ref{Fig.3}, only SDIC preserves the indentation on the cheeks. In the third row of Fig.~\ref{Fig.3}, SDIC and PTI successfully reconstruct the earrings and bangs. 
For the car domain, SDIC shows superior preservation of image details such as car lights, front, and reflective parts compared with other encoder-based methods. 
The above results show the effectiveness of SDIC. 

\subsection{Editing Results}
\noindent \textbf{Quantitative Evaluation.} There are no intuitive measures to evaluate the editing performance. Therefore, we calculate the $ID$ distance \cite{huang2020curricularface} between the original image and the manipulation one. Meanwhile, we conduct a user study to evaluate the editing results as done in  HFGI and CLCAE. Specifically, we collect 56 edited images of faces and cars for all the competing methods and ask 30 participants to choose the images with high fidelity and appropriate manipulation. The results are given in Table~\ref{tab1}. Our proposed method achieves the same $ID$ as PTI and greatly outperforms the other competing methods in terms of $User~Study$.

\begin{figure*} [!t]
	\centering 
	\includegraphics[width=0.9\textwidth]{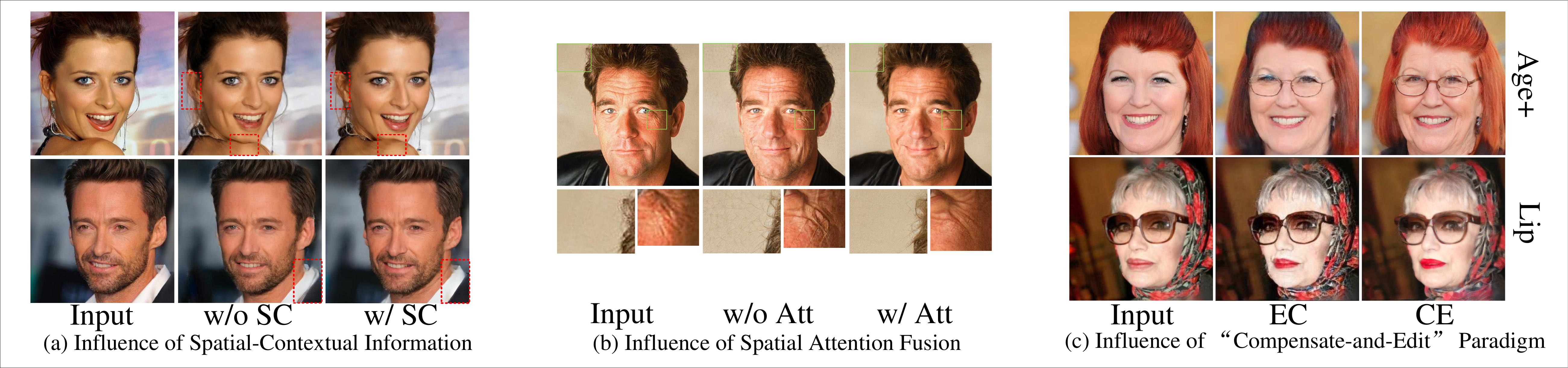} 
	
	\caption{{Ablation study results on the face domain}. In the figure, ``w/'' and ``w/o'' denotes ``with'' and ``without'', respectively. ``SC'' denotes ``Spatial-Contextual''. ``EC'' and ``CE'' denotes the ``Edit-and-Compensate'' paradigm and the ``Compensate-and-Edit'' paradigm, respectively. } 
\label{Fig.4} 
\end{figure*}

\noindent \textbf{Qualitative Evaluation.}
The image editing results are given in Fig.~3.
Compared with PTI, e4e, and Hyper, SDIC retains more detailed information in the editing results (e.g., the hat in the second row of Fig.~\ref{Fig.3}, the shadow and gap on the ground in the eighth row of Fig.~\ref{Fig.3}) while maintaining high image quality. In contrast, HFGI exhibits numerous artifacts (e.g., the neck in the second row and the right-side hair in the fourth row of Fig.~\ref{Fig.3}). FS loses  facial details and cannot edit car images well. CLCAE shows poor editing results (e.g., distortion in the neck in the fourth row of Fig.~\ref{Fig.3}) and small attributes changes  (e.g., minimal age reduction in the second row and almost no color change in the sixth row of Fig.~\ref{Fig.3}). 
In general, our method effectively balances fidelity and editability by incorporating spatial-contextual information.

\subsection{Ablation Studies}
\noindent \textbf{Influence of Spatial-Contextual Information.} 
We compare the reconstruction results obtained by our method with and without the two-branch spatial-contextual hourglass module. The results are given in Fig.~\ref{Fig.4}(a). For our method without the two-branch spatial-contextual hourglass module, we concatenate the two input images and directly feed them into the discrepancy map learning hourglass module. We select the two images under different conditions (i.e., side face and deformed mouth). 
Without the two-branch spatial-contextual hourglass module, our method 
shows substantial artifacts on {the edge of the portrait} and low fidelity for the mouth. In contrast, with the module, our method is more robust under different conditions, effectively removing artifacts and preserving more spatial details.

\noindent \textbf{Influence of Spatial Attention Fusion in DICN.} To compensate the latent map with the discrepancy information, we leverage the spatial attention fusion module. We evaluate the performance of our method with the spatial attention fusion module and with the conventional affine transformation.
The results are given in Fig.~\ref{Fig.4}(b).

Compared with our method with the spatial attention fusion module, our method with conventional affine transformation tends to give worse results. Some details of the generated image are overemphasized. For example, 
the wrinkles at the corners of the eyes are not natural while many reticulated lines appear in the background. 
This is because the affine transformation is only a linear transformation, leading to limited compensation results. In contrast, our model with the spatial attention fusion module adaptively selects informative parts 
and suppresses uninformative parts, generating high-quality images with a natural and smooth appearance.


\noindent \textbf{Influence of ``Compensate-and-Edit'' Paradigm.} 
We compare the ``compensate-and-edit'' and the ``edit-and-compensate'' paradigms. Specifically, we design a variant of our method based on the ``edit-and-compensate'' paradigm. That is, 
this variant predicts the discrepancy information between the initial edited and original images with DIPN and reconstructs the image with DICN. 
A comparison between this variant and our method is shown in Fig.~\ref{Fig.4}(c).

The variant gives worse editing results than our method. It considers the attribute changes as low-fidelity regions that do not match the original image. Thus, the network tends to correct them. Such a way reduces the effectiveness of the editability. In contrast, our method retains more detail after editing by the paradigm of ``first compensating and then editing''. 
The above experiments show the effectiveness of the ``compensate-and-edit'' paradigm. 

More ablation studies and applications of our method can refer to \textit{Supplement A and C}.


\section{Conclusion}
In this paper, we introduce a novel SDIC method, consisting of DIPN and DICN. Following the ``compensate-and-edit'' paradigm, SDIC first generates the spatial-contextual guided discrepancy information between the original image and the initial reconstructed image by DIPN and then 
compensates both the latent code and the GAN
generator with the discrepancy information by DICN.
Experimental results
show that our method can strike a good distortion-editability trade-off at a fast inference speed, which shows the effectiveness and efficiency of our method. 

One limitation of our proposed method is the
difficulty in handling large
manipulation cases (see \textit{Supplement D} for failure cases).
We intend to explicitly align the edited image and the original image in future work. 



\section{Acknowledgments}
This work was supported by the National Natural Science Foundation of China under Grants 62372388, 62071404, and U21A20514, and by the Open Research Projects of Zhejiang Lab under Grant 2021KG0AB02.

\bibliographystyle{plainnat}

\bibliography{aaai24}

\section{A. More Ablation Studies}

{
	\noindent \textbf{Influence of Spatial-Contextual Information and Spatial Attention Fusion.} We evaluate the influence of spatial contextual information and spatial attention fusion on the final performance. The results are given in Table ~\ref{tab1}. 
	Our SDIC achieves better results in terms of $L_2$, $LPIPS$, and $ID$ than our method without spatial attention fusion (i.e., `SDOC w/o Att'). This shows the importance of our spatial attention fusion. SDIC and `SDIC w/o Att' outperform `SDIC w/o SC \& Att`, indicating the necessity of exploiting the spatial-contextual information for GAN inversion.  
}

\noindent \textbf{Influence of the Layer of the Generator to Compensate.} In this paper, we leverage the spatial-contextual guided discrepancy map to compensate layer 7 of the GAN generator as suggested by HFGI. In this subsection, 
we evaluate the influence of the layer of the generator to compensate. We select layers 5, 7, and 9 of the generator for evaluation. 
The results are shown in Table ~\ref{tab2} and Fig.~\ref{Fig.9}.

From Table ~\ref{tab2}, when we compensate for the distortion information in the later layers, the fidelity of the generated images improves. However, the risk of overfitting also increases, leading to noticeable artifacts in the edited images (see Fig.~\ref{Fig.9}). 
To balance the distortion-editability trade-off, we choose layer 7 of the generator for information compensation in all our experiments.

\begin{figure*} [!t]
	\centering 
	\includegraphics[width=0.73\textwidth]{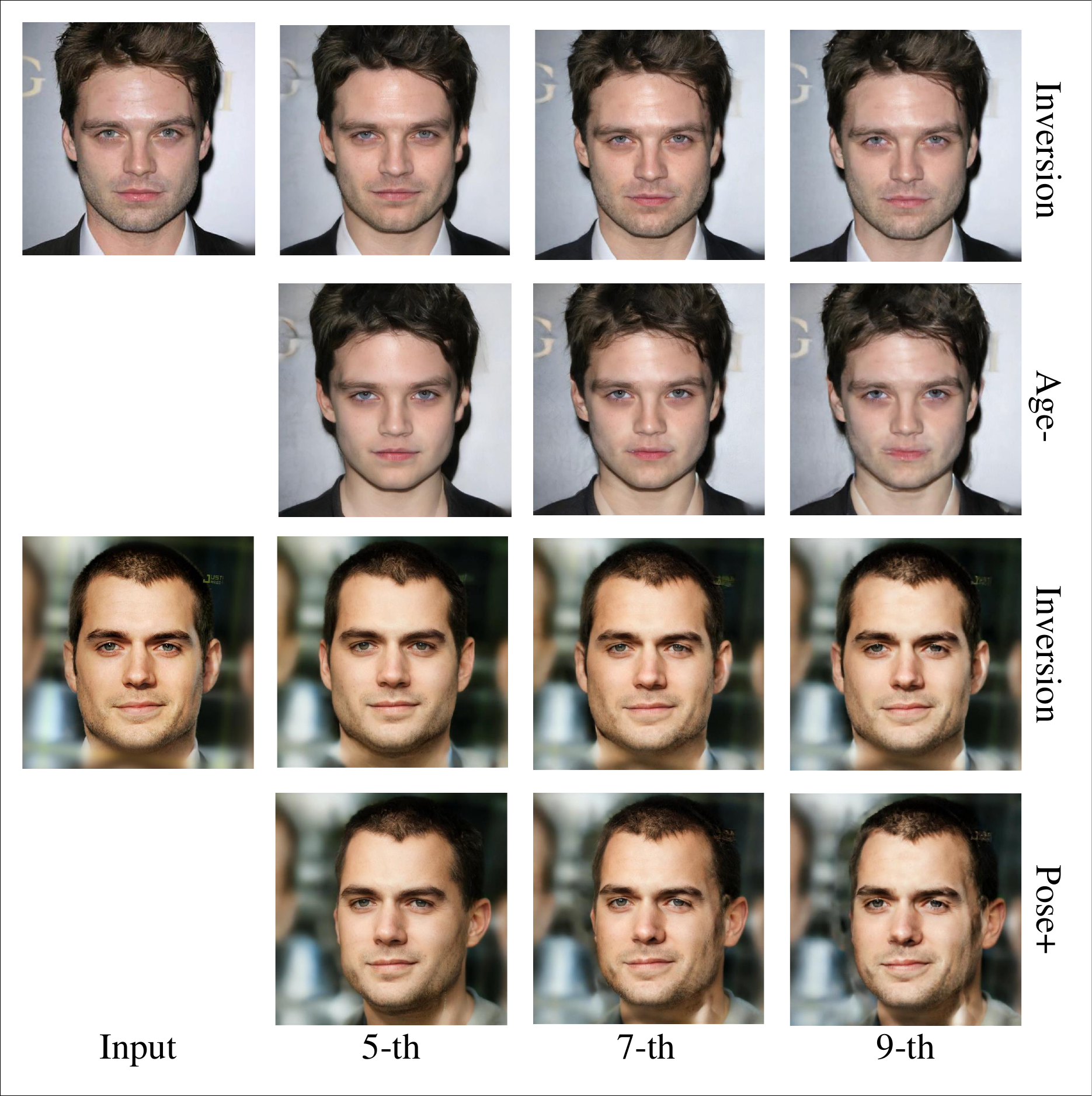} 
	
	\caption{Qualitative comparison for the influence of the layer of the generator to compensate on the face domain.} 
	\label{Fig.9} 
\end{figure*}

\section{B. More Visual Comparison Results}

We give more visual comparison results. 
\subsubsection{Face Domain.} In Fig.~\ref{Fig.5}, we show the inversion results and the editing results obtained by our SDIC, two high-fidelity GAN inversion methods (HFGI and CLCAE), and e4e (the pre-trained model used in SDIC) in the face domain. 
We can observe that e4e exhibits the excellent editability but lacks fidelity. HFGI and CLCAE greatly improve the fidelity of the reconstructed images at the cost of low editability. HFGI struggles to give good results with extreme deformations (e.g., pose+ in the third row of Fig.~\ref{Fig.5}). The editing results of CLCAE are not evident (e.g., adding a beard in the fourth row and increasing age in the seventh row of Fig.~\ref{Fig.5}). In general, SDIC achieves higher fidelity than HFGI and CLCAE and shows similar editability to e4e.


\subsubsection{Car Domain.} 
In Fig.~\ref{Fig.6}, we show the inversion results and the editing results obtained by our SDIC, HyperStyle, CLCAE, and e4e in the car domain. 
It can be observed that e4e and HyperStyle have better editability but lower fidelity compared with CLCAE and SDIC (e.g., the antenna in the first row, the bridge in the background in the sixth row, the cross on the front of the car in the seventh row, and the car headlights in the ninth row of Fig.~\ref{Fig.6}). CLCAE exhibits poor editability, generating numerous artifacts in the editing results and losing details from the original reconstructed image after editing (e.g., the antenna in the second row and the front of the car in the fourth row of Fig.~\ref{Fig.6}). In a word, SDIC demonstrates editability on par with HyperStyle and e4e while maintaining the details of the reconstructed image even after editing.

\begin{table}[!t]
	\begin{center}
		\tabcolsep=0.1cm
		\renewcommand\arraystretch{1}
		\begin{tabular}{l |c c c c } 
			\hline
			Method & $L_2\downarrow$ & $LPIPS\downarrow$ & $ID (ArcFace)\uparrow$ \\
			\hline
			SDIC w/o SC \& Att & 0.020 & 0.103 & 0.869  \\
			
			SDIC w/o Att & 0.009 & 0.062 & 0.929  \\
			
			SDIC & 0.007 & 0.057 & 0.948  \\
			\hline
		\end{tabular}
		
		\caption{Quantitative comparison for the influence of spatial-contextual information and spatial attention fusion.}
		\label{tab1}
	\end{center}
\end{table}

\begin{table}[!t]
	\begin{center}
		\renewcommand\arraystretch{1}
		\tabcolsep=0.33cm	\begin{tabular}{l |c c c c } 
			\hline
			Layer & $L_2\downarrow$ & $LPIPS\downarrow$ & $ID (ArcFace)\uparrow$ \\
			\hline
			5 & 0.176 & 0.090 & 0.838  \\
			
			7 & 0.007 & 0.057 & 0.948  \\
			
			9 & 0.004 & 0.039 & 0.972  \\
			\hline
		\end{tabular}
		
		\caption{Quantitative comparison for the influence of the layer of the
			generator to compensate on the face domain.}
		\label{tab2}
	\end{center}
\end{table}

\section{C. Additional Applications}
We apply our method to the text-to-image generation method StyleCLIP \cite{patashnik2021styleclip} and the style transfer method StyleGAN-NADA \cite{gal2022stylegan}, respectively. 
StyleCLIP combines StyleGAN \cite{karras2019style} and CLIP \cite{radford2021learning} to perform text-to-image synthesis. It allows the users to generate images by optimizing the latent code of a pre-trained StyleGAN model with text embeddings.  StyleGAN-NADA adjusts the generator to various domains with different styles and shape characteristics by using natural language prompts. It provides a flexible and efficient way to adapt the generative model to new contexts and generate diverse and high-quality images based on textual descriptions and prompts. 
The results are given in Fig.~\ref{Fig.8} and Fig.~\ref{Fig.7}. SDIC effectively improves the fidelity of the results compared with the original methods.


\section{D. Limitations }
In this paper, we leverage the spatial-contextual guided discrepancy map to compensate for the information loss during inversion. Our method achieves a good distortion-editability trade-off for the editing of most attributes. However, for some challenging cases (some failure cases are given in Fig.~\ref{Fig.10}), such as significant viewpoint changes, 
our method may give unsatisfactory editing results (generating artifacts). This is due to the extreme misalignment between the reconstructed image and the original image. 
Note that such a problem also exists in other GAN inversion methods (such as HIGI and CLCAE).

\newpage
\begin{figure*} [!t]
	\centering 
	\includegraphics[width=0.9\textwidth]{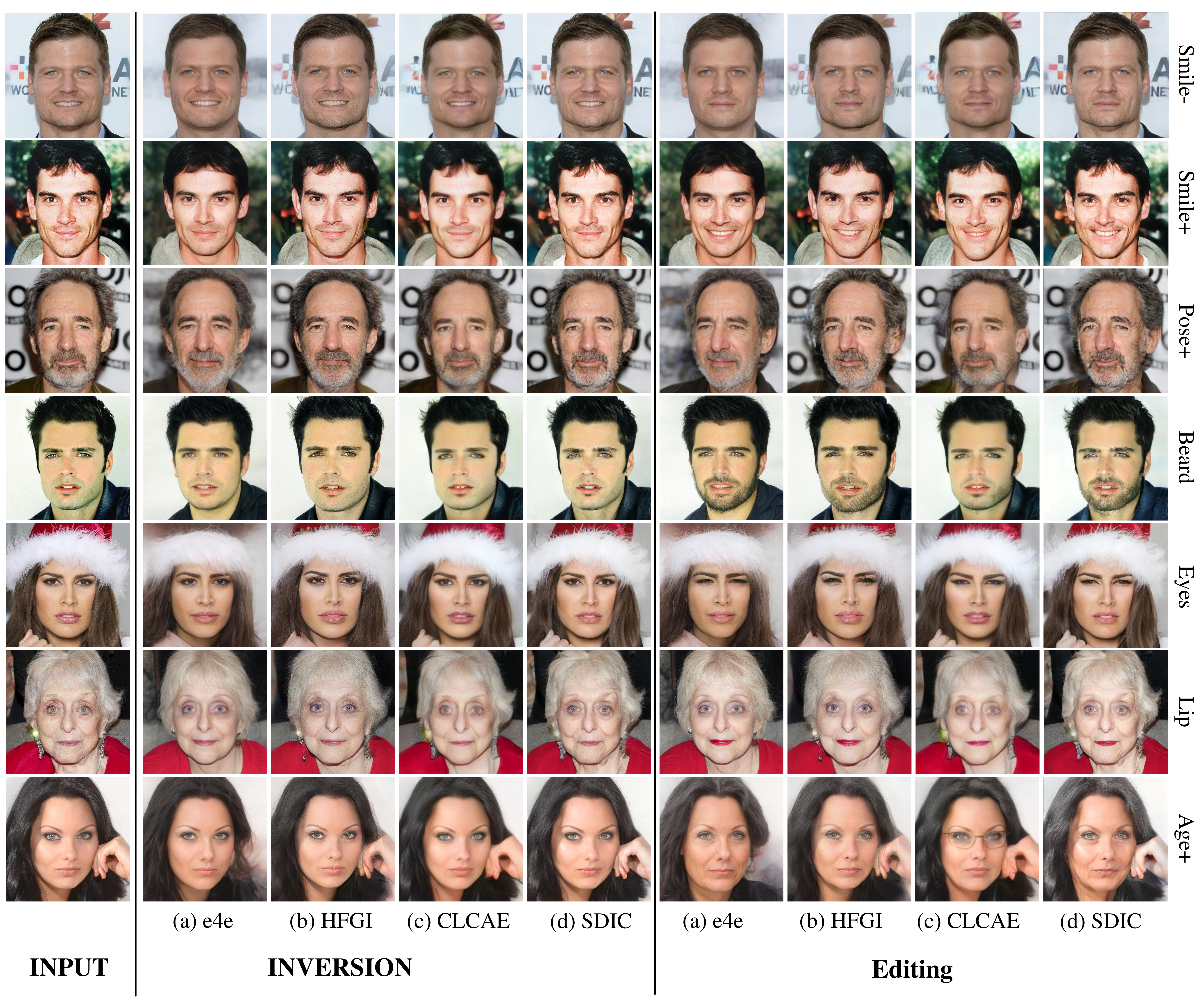} 
	
	\caption{Comparison in the face domain. } 
	\label{Fig.5} 
\end{figure*}

\begin{figure*} [!t]
	\centering 
	\includegraphics[width=0.9\textwidth]{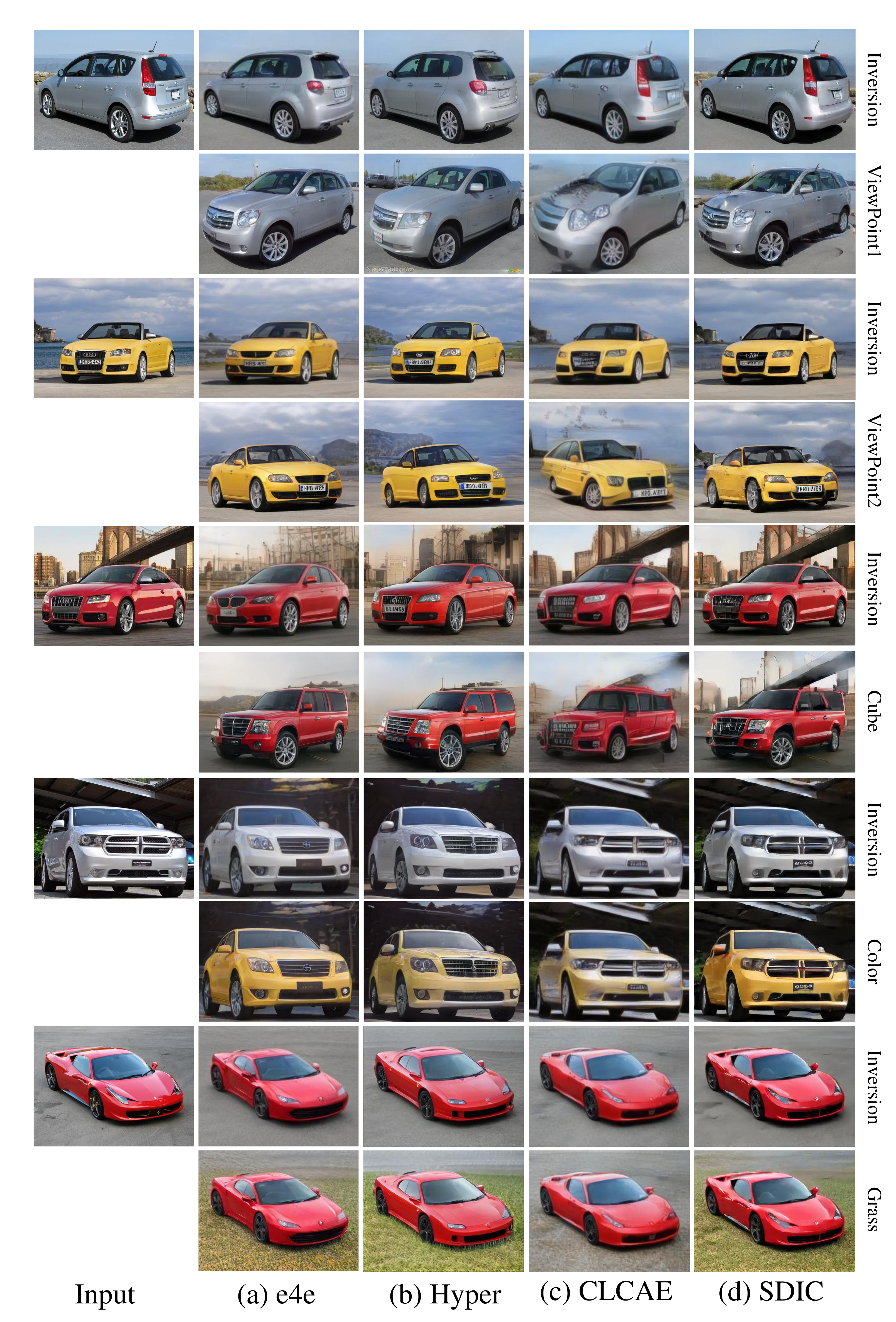} 
	
	\caption{Comparison in the car domain.} 
	\label{Fig.6} 
\end{figure*}

\begin{figure*} [!t]
	\centering 
	\includegraphics[width=1\textwidth]{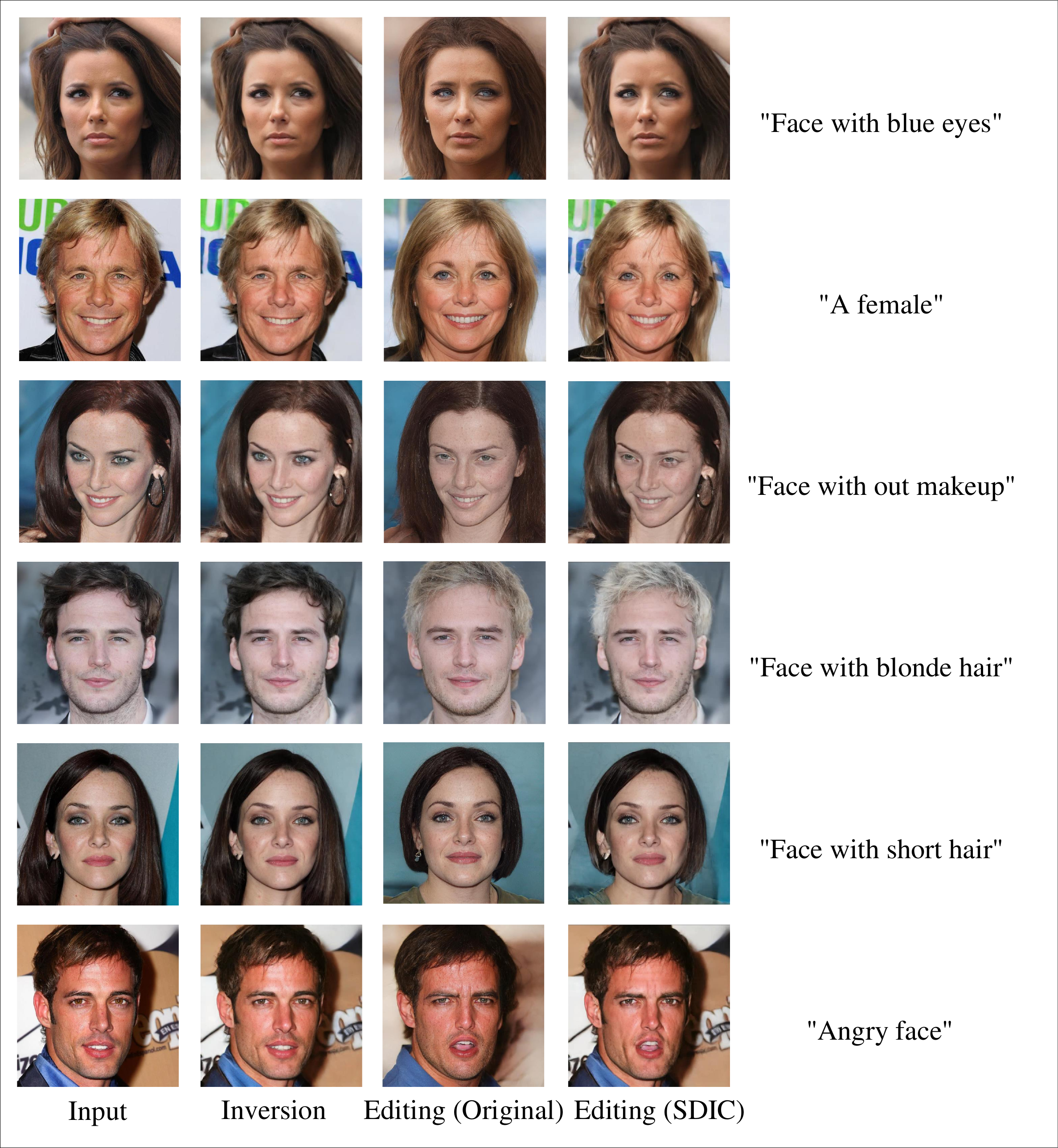} 
	
	\caption{Application in StyleClip.} 
	\label{Fig.8} 
\end{figure*}
\begin{figure*} [!t]
	\centering 
	\includegraphics[width=0.8\textwidth]{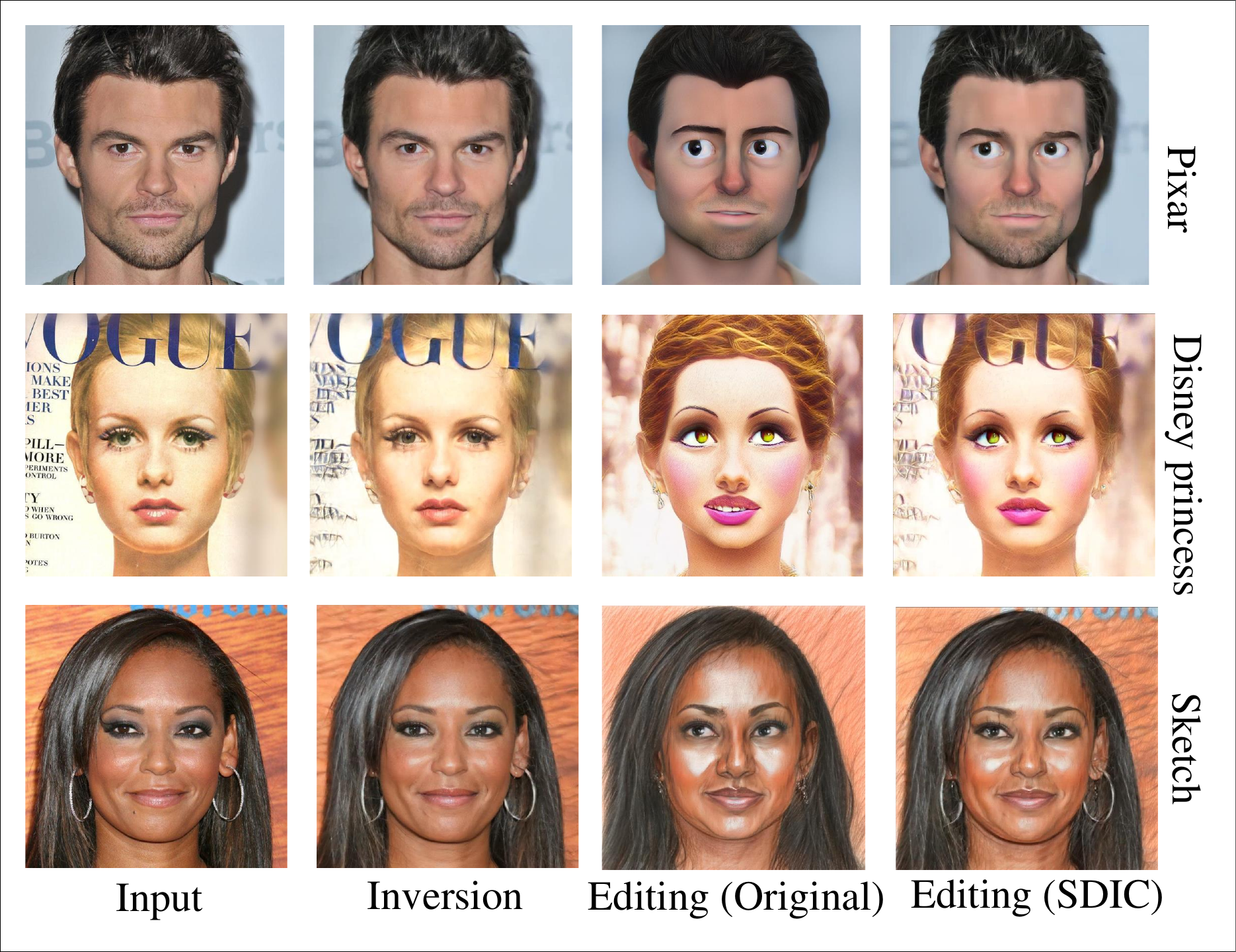} 
	
	\caption{Application in Styla-DANA.} 
	\label{Fig.7} 
\end{figure*}

\begin{figure*} [!t]
	\centering 
	\includegraphics[width=0.8\textwidth]{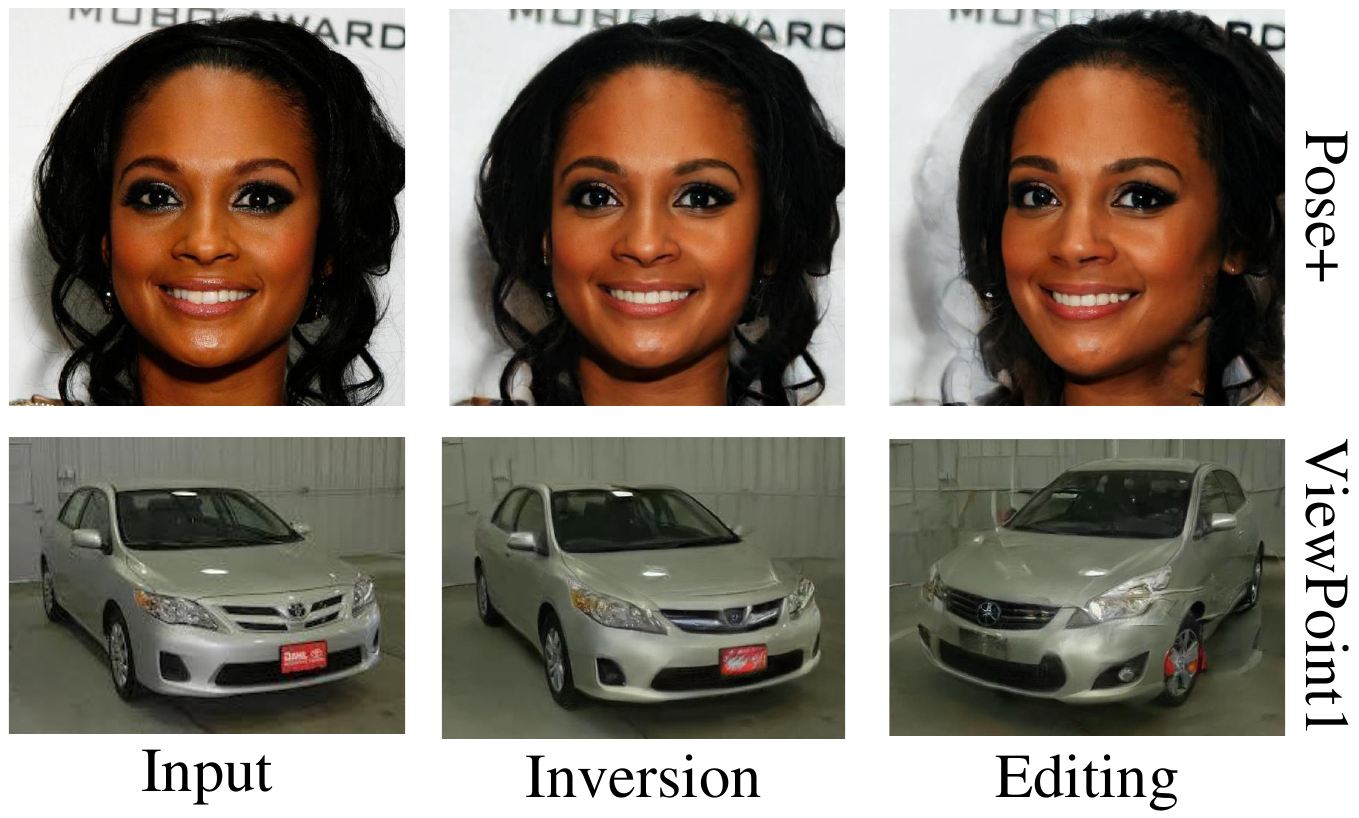} 
	
	\caption{Failure cases.} 
	\label{Fig.10} 
\end{figure*}

\end{document}